\documentclass[runningheads]{llncs}

\usepackage[T1]{fontenc}

\usepackage{graphicx,verbatim}

\usepackage{hyperref}
\usepackage{color}

\usepackage{float}
\usepackage{amssymb} 
\usepackage{mathtools}
\usepackage{graphicx}
\usepackage{verbatim}
\usepackage{tikz}                 
\usepackage{booktabs}             
\usepackage{multirow}        
\usepackage{esvect}         
\usepackage{cite}                 
\usepackage{array}                
\usepackage{float}
\usepackage{bm}

\begin{document}



\title{Label-Efficient Cross-Modality Generalization for Liver Segmentation in Multi-Phase MRI}

\author{
Quang-Khai Bui-Tran\inst{1}\thanks{Equal contribution}
\and
Minh-Toan Dinh\inst{2*}
\and
Thanh-Huy Nguyen\inst{3*}
\and \\
Ba-Thinh Lam\inst{1}
\and
Mai-Anh Vu\inst{4}
\and
Ulas Bagci\inst{5}\thanks{Corresponding author: \email{ulas.bagci@northwestern.edu}}
}

\authorrunning{Quang-Khai, Minh-Toan, Thanh-Huy et al.}

\institute{
Ho Chi Minh University of Science, Vietnam
\and
National Central University, Taiwan
\and
Carnegie Mellon University, PA, USA
\and
University of Houston, TX, USA
\and
Northwestern University, IL, USA
}

\maketitle              

\begin{abstract}
Accurate liver segmentation in multi-phase MRI is vital for liver fibrosis assessment, yet labeled data is often scarce and unevenly distributed across imaging modalities and vendor systems. We propose a \textit{label-efficient} segmentation approach that promotes \textit{cross-modality generalization} under real-world conditions, where GED4 hepatobiliary-phase annotations are limited, non-contrast sequences (T1WI, T2WI, DWI) are unlabeled, and spatial misalignment and missing phases are common.
Our method integrates a foundation-scale 3D segmentation backbone adapted via fine-tuning, co-training with cross pseudo supervision to leverage unlabeled volumes, and a standardized preprocessing pipeline. Without requiring spatial registration, the model learns to generalize across MRI phases and vendors, demonstrating robust segmentation performance in both labeled and unlabeled domains.
Our results exhibit the effectiveness of our proposed label-efficient baseline for liver segmentation in multi-phase, multi-vendor MRI and highlight the potential of combining foundation model adaptation with co-training for real-world clinical imaging tasks.


\end{abstract}

\section{Introduction}

Medical image segmentation has existed and been an important task for a long time, primarily due to its inherent interpretable nature, which is helpful to doctors in early diagnosis and treatment \cite{ma2024segment, azad2024medical, nguyen2023towards}. To date, there have been numerous medical image segmentation studies that were conducted on various organs of the human body, and one of them is the liver \cite{ma2024segment,zhu2024medical,isensee2024nnu,heller2023kits21}. Especially, under the label-scarce scenario, many noticeable works\cite{nguyen2025semi,pham2025fetal,nguyensemi} have tackled the partially annotated medical datasets and shown promising results on various settings.

The liver is one of the most important internal organs responsible for critical physiological functions. Additionally, the liver has complex anatomical structures \cite{juza2014clinical}. This complexity is further exacerbated by coupling with various pathological changes and high heterogeneity in medical images. One of the common liver diseases with such pathological changes today is liver fibrosis, a progressive condition resulting from chronic liver injury. Therefore, to diagnose effectively, it is crucial to segment the liver and accurately assess the fibrosis stage. 

In response to these clinical needs, the CARE (Comprehensive Analysis \& computing of REal-world medical images) 2025 challenge \cite{liu2025merit,gao2023reliable,wu2022meru}, in conjunction with MICCAI 2025, aims to advance real-world medical image analysis. In the CARE 2025 challenge, there are a total of five tracks that focus on distinct organs and their specific problems. In these tracks, CARE-Liver aims to advance the progress of Liver Fibrosis quantification and analysis methods. This track includes a crucial task, LiSeg - the automatic liver segmentation for liver MRI scans from multi-sequence and multi-center with limited ground truth.

The overall objective of the LiSeg task is to utilize the multi-modality MRI scans with the limitation of ground truth to segment the liver accurately. The challenges in this task are the limited liver mask annotations, along with the distribution shift due to the collection of data from multiple centers. Furthermore, the LiSeg task is divided into two sub-tasks corresponding to the segmentation of two different types of data: contrast and non-contrast data. However, only the ground truth for contrast data is provided. Therefore, the LiSeg task not only requires effective segmentation with limited ground truth data for a single data type, but also requires effective segmentation with unannotated non-contrast data. In view of the aforementioned considerations, we provide a comprehensive and standardized assessment of semi-supervised learning methods in this challenging task of liver segmentation.

In this paper, we propose a label-efficient liver segmentation framework tailored for multi-phase, multi-vendor MRI under limited annotation and cross-modality domain shift. To address the scarcity of annotated hepatobiliary-phase (GED4) data and the distributional gap across imaging centers, we leverage a suite of semi-supervised learning techniques, including Cross Pseudo Supervision (CPS)\cite{chen2021semi}, Bidirectional Copy-Paste (BCP)\cite{bai2023bidirectional}, and MiDSS\cite{ma2024constructing}, to exploit unlabeled GED4 and non-contrast sequences effectively. As our backbone, we adopt STU-Net\cite{huang2023stu}, a scalable and transferable segmentation model pretrained on large-scale multi-organ datasets, and further fine-tune it on the ATLAS liver segmentation dataset to improve adaptation to the target domain. This fine-tuning step provides a strong initialization prior to applying semi-supervised strategies. We validate our approach on the CARE-Liver challenge LiSeg task and conduct extensive ablation studies to analyze the contribution of each component in our pipeline, demonstrating strong generalization across MRI modalities and scanner vendors with minimal supervision.
\section{Methodology}
\subsection{Problem Definition}

The LiSeg Task of CARE-Liver 2025 challenge requires segmenting the liver in multi-phase fibrosis across multiple modalities, which specifically are categorized into two types: contrast (Gadolinium ethoxybenzyl diethylenetriamine pentaacetic acid enhanced fourth phase - GED4), and non-contrast data (T1-weighted imaging - T1WI, T2-weighted imaging - T2WI). The dataset of this challenge comprises several centers (Vendor A, B, and C), a total of 610 patients. All patients were diagnosed with liver fibrosis and underwent multi-phase MRI scans. However, only GED4 images of ten patients from each center are annotated, while other modalities and patients' counterparts are labeled.

Let $\mathsf{V}_n^{(p)} \in \mathbb{R}^{D \times W \times H}$ denote the volume for patient $n$ at phase $p \in \mathcal{P}$ and $\mathcal{P} = \{\text{GED4}, \text{T1WI}, \text{T2WI}\}$. Formally, the training labeled set is defined as $\mathcal{D}^l = \{(\mathsf{V}_n^{(\text{GED4})}, \mathsf{G}_n)\}_{n=1}^{N}$ where $\mathsf{G}_n \in \{0,1\}^{D \times W \times H}$ represents the binary liver segmentation mask and $N = 30$ (10 annotations per center). The training unlabeled set comprises $\mathcal{D}^u = \{\mathsf{V}_m^{(p)}\}_{m=1, p \in \mathcal{P}}^{M}$ where $M$ represents the vast majority of patients whose unlabeled volumes across all phases and $M = 330$.

The primary objective is to learn two distinct segmentation functions corresponding two types of data, contrast and non-contrast: $f_\theta^{(\text{contrast})}: \mathbb{R}^{D \times W \times H} \rightarrow [0,1]^{D \times W \times H}$ for GED4 volumes and $f_\phi^{(\text{non-contrast})}: \mathbb{R}^{D \times W \times H} \rightarrow [0,1]^{D \times W \times H}$ for T1WI, and T2WI phases. While the challenge in GED4 images training is excessively limited labels, other modalities' counterparts do not have any direct supervision, requiring knowledge transfer from the limited GED4 annotations.

\subsection{Scalable and Transferable Medical Image Segmentation Models}

Although nnU-Net \cite{isensee2021nnu} has demonstrated strong performance across various tasks by automatically adapting to dataset-specific properties (e.g., input patch size, spacing), its reliance on dynamic hyperparameter selection limits the transferability of pretrained models. Core architectural parameters such as the number of resolution stages, kernel sizes, and downsampling/upsampling ratios are derived from dataset heuristics, making nnU-Net models less generalizable across tasks without retraining.

To address this, STU-Net \cite{huang2023stu} introduces a scalable and transferable framework by categorizing hyperparameters into two types: (i) \emph{weight-related} (e.g., resolution stages), and (ii) \emph{weight-unrelated} (e.g., input patch size). The former are fixed (e.g., six resolution stages and $(3\times3\times3)$ convolution kernels), ensuring consistent model weights across tasks, while the latter follow nnU-Net's robust defaults.

STU-Net further incorporates several architectural improvements: 
\textbf{(1) Residual blocks} replace basic Conv-IN-LeakyReLU blocks in the encoder and decoder, using two $3\times3\times3$ convolutions with a skip connection to stabilize deep training and preserve gradient flow. 
\textbf{(2) Downsampling blocks} employ a dual-branch design: a main branch with two convolutions (stride 1 then 2) and a skip branch with a $1\times1\times1$ convolution (stride 2), and their outputs are summed to retain spatial features. 
\textbf{(3) Upsampling blocks} use nearest-neighbor interpolation followed by a $1\times1\times1$ convolution, avoiding learnable parameters and improving transferability across resolutions. 
\textbf{(4) Compound scaling}, inspired by EfficientNet \cite{tan2019rethinking}, jointly increases network depth and width at each resolution stage, enhancing both receptive field and representational capacity.

Together, these design choices make pretrained STU-Net a more generalizable and weight-compatible backbone for transfer across diverse medical tasks.

\begin{table}[ht]
\centering
\caption{STU-Net configurations at different scales.}
\begin{tabular}{|c|c|c|c|c|}
\hline
\textbf{Model} & \textbf{Depth} & \textbf{Width} & \textbf{Parameters (M)} & \textbf{FLOPs (T)} \\ \hline
STU-Net-S & (1,1,1,1,1,1) & (16,32,64,128,256,256) & 14.60 & 0.13 \\ \hline
STU-Net-B & (1,1,1,1,1,1) & (32,64,128,256,512,512) & 58.26 & 0.51 \\ \hline
STU-Net-L & (2,2,2,2,2,2) & (64,128,256,512,1024,1024) & 440.30 & 3.81 \\ \hline
\end{tabular}
\label{tab:stunet_scales}
\end{table}

\textbf{Large-Scale Supervised Pretraining:}  
STU-Net is pretrained on the \emph{TotalSegmentator} dataset, which includes 104 anatomical classes. The pretraining is conducted for 4000 epochs, which is four times longer than standard nnU-Net training and incorporates mirror data augmentation to enhance generalization for downstream tasks.

\subsection{Cross Pseudo Supervision for 3D Segmentation}
\begin{figure}[ht]
    \centering
    \includegraphics[width=1.0\textwidth]{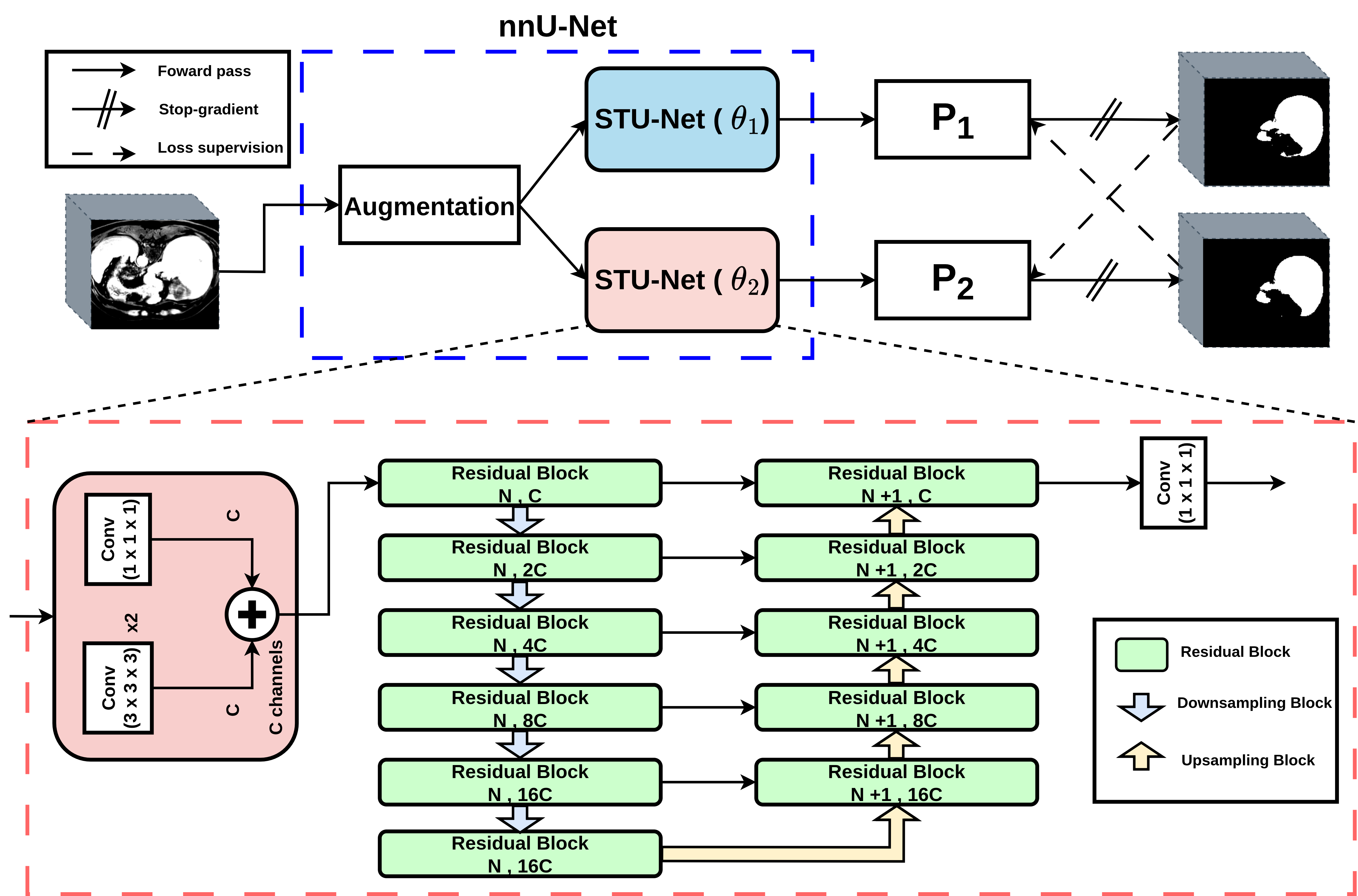}
    \caption{\textbf{Overview of the semi-supervised segmentation cross pseudo supervision framework.} The framework utilizes the auto configurations of nnU-Net for loading data and augmentation, while the backbone uses the STU-Net model with a strong pretrained weight. Then deploy cross pseudo supervised between the two models} 
    \label{fig:overview} 
\end{figure}   

We utilize a dual-network training strategy based on the \textit{STU-Net} architecture,  which is a strong foundation model and suitable for various tasks and data.  The framework consists of two identical 3D segmentation networks with independent parameters:

\begin{align}
\mathsf{P}_1 & = \mathrm{STU\mbox{-}Net}(\mathsf{V}; \boldsymbol{\theta}_1), \\
\mathsf{P}_2 & = \mathrm{STU\mbox{-}Net}(\mathsf{V}; \boldsymbol{\theta}_2),
\end{align}

where $\mathsf{P}_1, \mathsf{P}_2 \in \mathbb{R}^{ C \times D \times W \times H}$ are voxel-wise confidence maps after softmax normalization over $C$ classes. 

The same augmented input volume $\mathsf{V}$ is provided to both networks to encourage consistent learning while enabling complementary supervision. To simplify notation, the forward path can be expressed as:

\begin{align}
\mathsf{V} \rightarrow \mathrm{STU\mbox{-}Net}(\boldsymbol{\theta}_1) \rightarrow \mathsf{P}_1 \rightarrow \mathsf{Y}_1, \notag\\
\phantom{\mathsf{V}} \quad\quad\quad\searrow \mathrm{STU\mbox{-}Net}(\boldsymbol{\theta}_2) \rightarrow \mathsf{P}_2 \rightarrow \mathsf{Y}_2,
\end{align}
where $\mathsf{Y}_1$ and $\mathsf{Y}_2$ are pseudo segmentation maps obtained by applying the $\mathrm{argmax}$ operation to $\mathsf{P}_1$ and $\mathsf{P}_2$, respectively. Each pseudo label $\mathbf{y}_{1i}$ ($\mathbf{y}_{2i}$) is a one-hot vector corresponding to the predicted class at voxel $i$.

The total training loss consists of two components:  
(i) a \textbf{supervised loss} $\mathcal{L}_s$ applied to labeled data, and  (ii) a \textbf{cross pseudo supervision loss} $\mathcal{L}_{cps}$ applied to both labeled and unlabeled data.

\noindent \textbf{Supervised Loss:} For labeled volumes, we employ the standard voxel-wise cross-entropy loss for each network:
\begin{align}
\mathcal{L}_s &= \frac{1}{|\mathcal{D}^l|} \sum_{\mathsf{V} \in \mathcal{D}^l}
\frac{1}{W \times H \times D} 
\sum_{i=0}^{W \times H \times D} 
\Big( \ell_{ce}(\mathbf{p}_{1i}, \mathbf{y}^*_{1i}) \notag\\
&\quad\quad + \ell_{ce}(\mathbf{p}_{2i}, \mathbf{y}^*_{2i}) \Big),
\end{align}
where $\ell_{ce}$ denotes the cross-entropy function, 
$\mathbf{p}_{1i}$ and $\mathbf{p}_{2i}$ are the softmax probability vectors for voxel $i$, 
and $\mathbf{y}^*_{1i}$, $\mathbf{y}^*_{2i}$ are the corresponding ground truth one-hot vectors.

\noindent \textbf{Cross Pseudo Supervision Loss:} In the unlabeled case, pseudo labels are generated from one network and used to supervise the other.  This bidirectional supervision is defined as:
\begin{align}
\mathcal{L}_{cps}^u &= \frac{1}{|\mathcal{D}^u|} \sum_{\mathsf{V} \in \mathcal{D}^u}
\frac{1}{W \times H \times D} 
\sum_{i=0}^{W \times H \times D} 
\Big( \ell_{ce}(\mathbf{p}_{1i}, \mathbf{y}_{2i}) \notag\\
&\quad\quad + \ell_{ce}(\mathbf{p}_{2i}, \mathbf{y}_{1i}) \Big),
\end{align}
where $\mathbf{y}_{1i}$ and $\mathbf{y}_{2i}$ are pseudo labels produced by the other network.

Similarly, we define $\mathcal{L}_{cps}^l$ for labeled volumes using the same formulation. The total cross pseudo supervision loss is a combination of the losses on both the labeled and unlabeled data, which is:
\begin{equation}
\mathcal{L}_{cps} = \mathcal{L}_{cps}^l + \mathcal{L}_{cps}^u.
\end{equation}

\subsubsection{Final Objective:}

The complete loss function is:
\begin{equation}
\mathcal{L} = \mathcal{L}_s + \lambda \mathcal{L}_{cps},
\end{equation}
where $\lambda$ is a balancing hyperparameter controlling the contribution of cross pseudo supervision, which allows the networks to improve their predictions mutually, leveraging information from both labeled and unlabeled 3D volumes.

\section{Experiments}

\subsection{Dataset}

\textbf{LiQA Dataset} The dataset comprises a multi-center cohort of 610 patients diagnosed with liver fibrosis, including 170 new cases beyond the CARE2024 challenge \cite{liu2025merit,gao2023reliable,wu2022meru}. Patients were scanned across four clinical centers using three MRI vendors \textit{Philips Ingenia 3.0T}, \textit{Siemens Skyra 3.0T}, and \textit{Siemens Aera 1.5T}, resulting in realistic multi-vendor variability. Each patient underwent multi-phase MRI, including T2-weighted (T2WI), diffusion-weighted (DWI), T1-weighted (T1WI), and Gd-EOB-DTPA–enhanced dynamic imaging at four phases: arterial (GED1, 25s post-injection), portal venous (GED2, 1min), delayed (GED3, 4min), and hepatobiliary (GED4, 20min). The training set includes 360 cases (vendor A: 130, B1: 170, B2: 60), but only 30 GED4 scans (8.3\%) have segmentation labels, posing a strong annotation scarcity challenge. Consequently, \textit{subtask 1} (T1WI, T2WI, DWI) operates fully unsupervised, while \textit{subtask 2} (GED4) requires label-efficient or semi-supervised strategies. The validation set contains 60 fully annotated cases (20 per vendor A, B1, B2). The test set includes 190 cases from all vendors (A: 40, B1: 40, B2: 40, C: 70), with vendor C held out to assess generalization. Key challenges include missing sequences (except GED4), inter-vendor protocol differences, and the lack of spatial pre-alignment, reflecting the heterogeneity of clinical liver MRI data.

\noindent \textbf{ATLAS Dataset} The ATLAS dataset from the MICCAI 2023 ATLAS challenge \cite{quinton2023tumour}, includes 60 public and 30 private patient cases. Each case contains multiple NIFTI-format MRI scans and corresponding labels. Images were acquired using various MRI machines and sequences, with one of three post-contrast phases (arterial, portal, or delayed) selected. Liver and tumour masks were merged into a single foreground for the accurate liver segmentation task.

\begin{table}[ht]
\centering
\caption{Sub-Task 1 – GED4 segmentation results on the private validation set from the organizers with STU-Net-S backbone. 
$\uparrow$: higher is better, $\downarrow$: lower is better. 
Bold indicates the best result, underline indicates the second best.}
\label{tab:ged4_results}
\begin{tabular}{|l|c|c|}
\hline
\textbf{Method} & \textbf{DSC} $\uparrow$ & \textbf{HD} $\downarrow$ \\
\hline
CPS& \textbf{0.9685} & \textbf{25.56} \\
\hline
BCP & \underline{0.9553} & \underline{86.69} \\
\hline
MiDSS & 0.9142 & 126.17 \\
\hline
\end{tabular}
\end{table}

\subsection{Experimental Setup}
Due to computational constraints, only 3D STU-Net-S is adopted as the backbone network initialized with pretrained weights and fine-tuned on the ATLAS dataset to adapt the foundation model to the task-specific setting. MRI data are preprocessed and augmented using the nnU-Net pipeline, including z-score normalization, random spatial transformations, and other standard augmentations. Adam optimizer is employed for all training. Fine-tuning on ATLAS is performed for 100 epochs with a learning rate of $1\times10^{-2}$ and a batch size of 2. Semi-supervised model training is carried out with a learning rate of $1\times10^{-3}$ for a maximum of 36 epochs and a batch size of 4. All experiments are implemented in PyTorch and executed on a single NVIDIA GeForce RTX 3090~Ti GPU.

\textbf{Evaluation Metrics:} For the LiSeg task, liver segmentation performance of our methods is assessed by using the Dice Similarity Coefficient (DSC) and Hausdorff Distance (HD), in accordance with the evaluation criteria of the challenge.

\subsection{Results}

\textbf{Sub-Task 1: Contrast-Enhanced Liver Segmentation (GED4)} The table~\ref{tab:ged4_results} shows the results on contrast-enhanced hepatobiliary phase (GED4) using the private validation set. Three methods compared include the state-of-the-art semi-supervised model CPS \cite{chen2021semi}, BCP \cite{bai2023bidirectional}, and MiDSS \cite{ma2024constructing}. Specifically, CPS achieves superior performance with a DSC of 0.9685 and an HD of 25.56. Both BCP and MiDSS, which adopt teacher–student architectures, achieved competitive results but still underperform compared to CPS.

\textbf{Sub-Task 2: Non-Contrast Liver Segmentation}
Table~\ref{tab:modality_results} reports segmentation results on non-contrast modalities without annotations. For T1WI, CPS with STU-Net-S fine-tuned on ATLAS achieves the best performance (DSC: 0.9465, HD: 62.84), outperforming the non-fine-tuned counterpart (DSC: 0.9392, HD: 89.90). Results on T2WI are notably lower (DSC: 0.7673, HD: 81.76), highlighting the additional difficulty of segmenting this modality in the absence of annotations. This can be explained by two different groups of reasons, including the difference in characteristics between the modality and our training scenario. 

Regarding the different features, physically, T2WI exhibits completely reversed signal characteristics compared to GED4 and T1WI. Specifically, while the liver appears hyperintense (bright) on GED4 due to hepatocytes' uptake of Gd-EOB-DTPA intermediate-to-high signal on T1WI, the liver appears hypointense (dark) on T2WI \cite{curvo2010hypointense}. This contrast pattern reversal creates a fundamental challenge for CNN feature transfer \cite{kang2024target}. CNN, which learns from 30 annotated GED4 images, tends to detect bright-to-dark transitions such as liver-to-background. In contrast, when it is applied to T2WI with dark liver regions, its features fail catastrophically with significant performance drops for the cross-modality medical image segmentation task.

In terms of training scenarios, despite our efforts to address label scarcity for cross-modality medical images, it still cannot completely address the domain gap problem. The external dataset used for pretraining was ATLAS, which included only T1 CE-MRI and no T2, resulting in a significant bias toward T1/contrast-enhanced patterns. A pseudo-labeling strategy with 220 cases was generated only for unlabeled GED4, with no direct supervision signal for the T2WI domain. As a result, our models have no chance to self-correct on the T2 distribution.

\begin{table}[ht]
\centering
\caption{Sub-Task 2 – Segmentation results on non-contrast modalities (T1WI, T2WI) from the private validation set provided by the organizers.}
\label{tab:modality_results}
\begin{tabular}{|l|l|c|c|}
\hline
\textbf{Modality} & \textbf{Method} & \textbf{DSC} $\uparrow$ & \textbf{HD} $\downarrow$ \\
\hline
\multirow{3}{*}{T1WI} 
& CPS with STU-Net-S (fine-tuned on ATLAS) & \textbf{0.9465} & \textbf{62.84} \\
\cline{2-4}
& CPS with STU-Net-S & 0.9392 & 89.90 \\
\cline{2-4}
& MiDSS with STU-Net-S + nnU-Net preprocessing & 0.8948 & 128.93 \\
\hline
T2WI 
& CPS with STU-Net-S (fine-tuned on ATLAS) & 0.7673 & 81.76 \\
\hline
\end{tabular}
\end{table}

\textbf{Ablation Study.}  
Table~\ref{tab:ged4_ablation} highlights the impact of foundation model pretraining and task-specific fine-tuning. The baseline CPS with a 3D U-Net backbone performs poorly (DSC: 0.7752, HD: 141.07), highlighting the limitations of training from scratch, particularly in semi-supervised settings with a large amount of unlabeled data. Replacing the backbone with STU-Net-S substantially improves performance (DSC: 0.9685, HD: 25.56). Further initializing the model with ATLAS fine-tuning weights yields the best results.

\begin{table}[ht]
\centering
\caption{Ablation study on Sub-Task 1. Comparison of different backbones, ATLAS fine-tuning, and augmentation strategies on the private validation set.}
\label{tab:ged4_ablation}
\begin{tabular}{|l|c|c|}
\hline
\textbf{Method} & \textbf{DSC} $\uparrow$ & \textbf{HD} $\downarrow$ \\
\hline
CPS with 3D U-Net & 0.7752 & 141.07 \\
\hline
CPS with STU-Net-S & \underline{0.9685} & \underline{25.56} \\
\hline
CPS with STU-Net-S (fine-tuned on ATLAS) & \textbf{0.9705} & \textbf{19.55} \\
\hline
MiDSS with STU-Net-S + nnU-Net preprocessing & 0.9578 & 37.85 \\
\hline
MiDSS with STU-Net-S + full augmentation & 0.9142 & 126.17 \\
\hline
MiDSS with STU-Net-S + weak augmentation & 0.8665 & 130.04 \\
\hline
\end{tabular}
\end{table}

Different processing pipelines are also employed to investigate their contributions to the overall performance. Following the MiDSS pipeline, full augmentation substantially degrades performance (DSC: 0.9142, HD: 126.17), while weak augmentation by disabling Elastic Transform, Random Crop \& Rotate, and Random Crop Zoom leads to an even larger drop (DSC: 0.8665, HD: 130.04). In contrast, adopting nnU-Net preprocessing with its automatically configured data loader achieves competitive results (DSC: 0.9578, HD: 37.85), highlighting the robustness of the nnU-Net pipeline compared to manual augmentation design.

\begin{table}[ht]
\centering
\caption{LiSeg evaluation results across different MRI modalities for liver segmentation. Results are reported for In-Domain (ID) and Out-of-Domain (OOD) test sets using Dice Similarity Coefficient (DSC) and Hausdorff Distance (HD) metrics.}
\label{tab:private_results}
\begin{tabular}{|l|c|c|c|c|}
\hline
\multirow{2}{*}{\textbf{Modality}} & \multicolumn{2}{c|}{\textbf{ID Test}} & \multicolumn{2}{c|}{\textbf{OOD Test}} \\
\cline{2-5}
 & \textbf{DSC} $\uparrow$ & \textbf{HD} $\downarrow$ & \textbf{DSC} $\uparrow$ & \textbf{HD} $\downarrow$ \\
\hline
GED4 & 94.56 & 56.59 & 97.26 & 21.52 \\
\hline
T1   & 86.42 & 163.91 & 73.51 & 102.79 \\
\hline
\end{tabular}
\end{table}

\subsection{Discussion}

\begin{figure}[ht]
\centering
\includegraphics[width=0.9\textwidth]{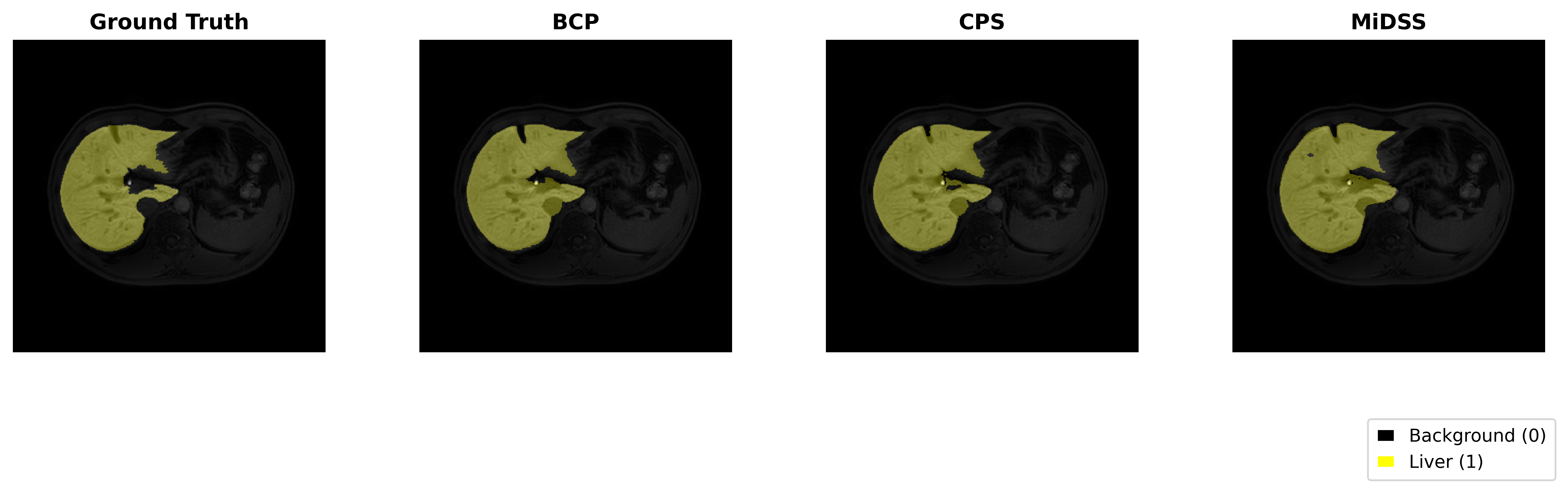}
\caption{Qualitative results on GED4 using different semi-supervised methods.}
\label{fig:plot_table_04}
\end{figure}

\begin{figure}[ht]
\centering
\includegraphics[width=0.9\textwidth]{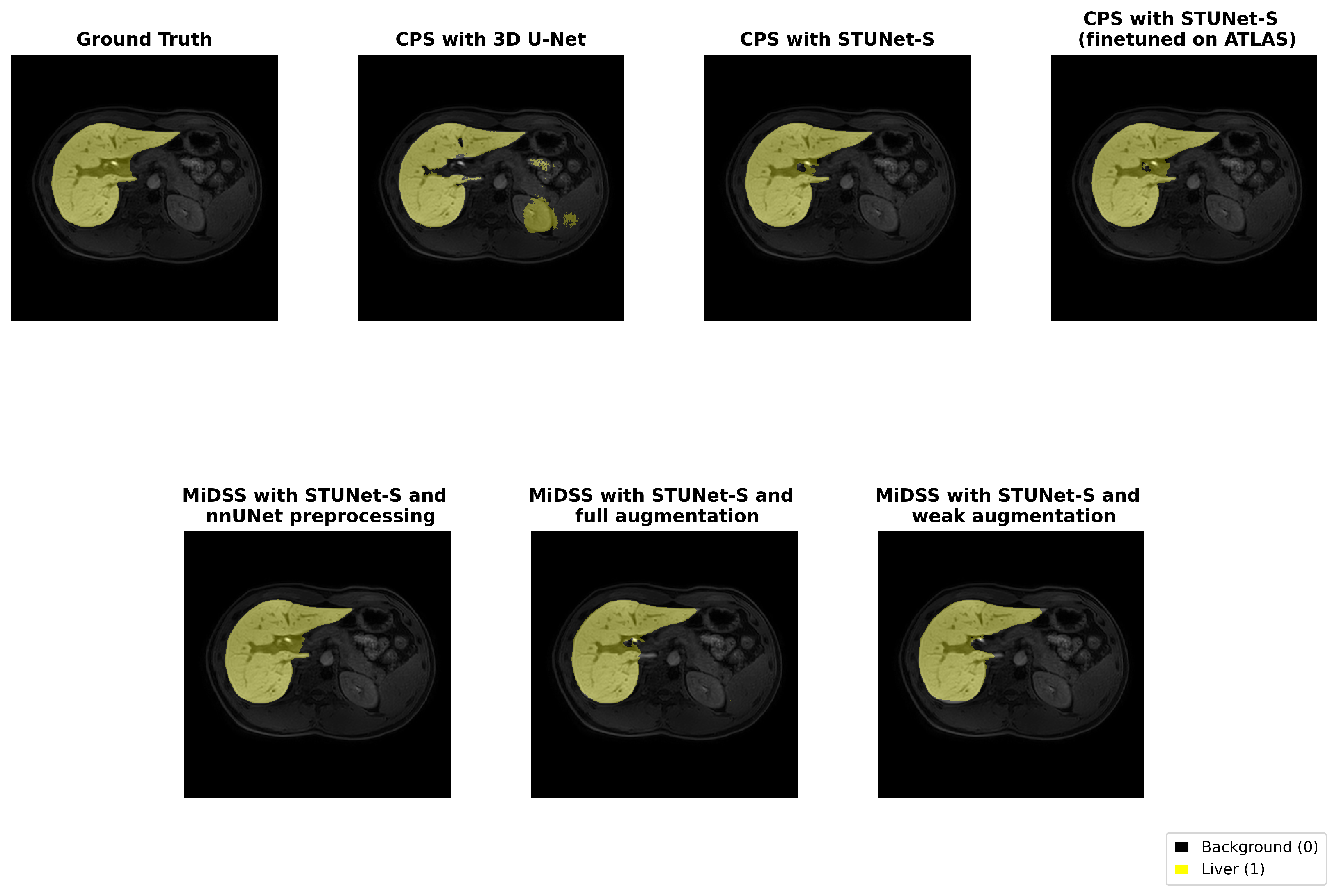}
\caption{Qualitative results on GED4 under different pretraining weights and preprocessing pipelines.}
\label{fig:plot_table_02}
\end{figure}

To clarify the performance gap between the methods, the limited annotated GED4 data is split into 5 cases for validation and the remaining cases for training. The qualitative results shown in Figures~\ref{fig:plot_table_04} demonstrate that the CPS method closely follows the liver boundaries in the ground truth. Additionally, Figure~\ref{fig:plot_table_02} further illustrates the benefits of fine-tuning the foundation model on ATLAS and leveraging nnU-Net preprocessing.

Our models are designed to address multiple subtasks concurrently. As reported in Table~\ref{tab:modality_results}, Table~\ref{tab:ged4_results}, and Table~\ref{tab:private_results}, performance on non-contrast modalities lags behind that of GED4, likely due to the scarcity of annotated data for these modalities. In future work, strategies that enhance cross-modality generalization will be investigated, with a particular focus on improving performance for non-contrast data.

Our models are also evaluated on a private dataset from the CARE-Liver challenge to validate the robustness, and their results are shown in Table~\ref{tab:private_results}. These results suggest that the models generalize well to out-of-distribution data, supporting their potential for use in clinical practice.  

\section{Conclusion}

In this work, we present a comprehensive study on liver MRI segmentation under the extreme annotation scarcity and multi-vendor heterogeneity of the LiQA dataset. By combining nnU-Net preprocessing, the STU-Net backbone with foundation model fine-tuning from ATLAS, and carefully designed semi-supervised strategies, we demonstrate substantial improvements in both contrast-enhanced and non-contrast settings. Our results on the LiQA dataset demonstrate the strength of the baseline, which integrates foundation model fine-tuning with semi-supervised learning for robust liver segmentation in real-world clinical settings.

\begin{credits}
\section{\ackname}

This research is partially supported by the following NIH grants: R01-HL171376 and U01-CA268808. We thank AI VIETNAM for supporting GPUs for training experiments.

\end{credits}

\bibliographystyle{splncs04}
\bibliography{references/references}

\end{document}